\RequirePackage{amsmath}
\documentclass[runningheads]{llncs}

\usepackage{graphicx}

\usepackage{graphicx} 
\usepackage{float}
\usepackage{pgf}
\usepackage{amssymb} 
\usepackage{url} 
\usepackage{mathtools} 
\usepackage{wrapfig}
\usepackage{float}
\usepackage[T1]{fontenc}
\usepackage{mwe}
\usepackage[caption=false]{subfig}
\usepackage{multirow}

\begin{document}
\title{Learn-able parameter guided Activation Functions}
%
%
\author{S.Balaji \inst{1}\orcidID{1} \and
T.Kavya\inst{1}\orcidID{2} \and
Natasha Sebastian\inst{2}\orcidID{2}}
\authorrunning{S.Balaji et al.}
%
\institute{HCL Technologies \and
Delhi Technological University
}
\maketitle
\begin{abstract}
In this paper, we explore the concept of adding learn-able slope and mean shift parameters to an activation function to improve the total response region. The characteristics of an activation function depend highly on the value of parameters. Making the parameters learn-able, makes the activation function more dynamic and capable to adapt as per the requirements of it's neighboring layers. The introduced slope parameter is independent of other parameters in the activation function. The concept was applied to ReLU to develop Dual Line and Dual Parametric ReLU activation function. Evaluation on MNIST and CIFAR10 show that the proposed activation function Dual Line achieves top-5 position for mean accuracy among 43 activation functions tested with LENET4, LENET5 and WideResNet architectures. This is the first time more than 40 activation functions were analyzed on MNIST and CIFAR10 dataset at the same time. The study on the distribution of positive slope parameter $\beta$ indicates that the activation function adapts as per the requirements of the neighboring layers. The study shows that model performance increases with the proposed activation functions.

\end{abstract}

\section{Introduction}
The activation functions used across the layers of deep neural networks play a significant role in the ability of the whole network to achieve good performance. Though each layer of a deep neural network can have different requirements, the convention is to use the same activation function at each output of a layer. Therefore using a good activation function suitable for every output node in the layer is essential. \\
Rectified Linear Units (ReLU) \cite{47} and its variants such as Leaky ReLU \cite{2}, Parametric ReLU (PReLU)\cite{9} are the most commonly used activation functions due to their simplicity and computational efficiency. The introduction of learn-able parameter in the negative axis for PReLU increased the overall response region. Though extra parameters are introduced in PReLU activation function, the number of parameters added due to PReLU is negligible compared to the total number of parameters in the network. Another activation function that makes use of learn-able parameters is Parametric ELU(PELU) \cite{8}. Learn-able parameters in PELU activation function adopted characteristics as per the requirements of the training stage. In the case of PELU, the positive axis parameter is dependent as it is defined as the ratio of parameters used to alter exponential decay and saturation point. In this paper, we introduce two activation functions, with a positive slope parameter which is independent of other parameters, allowing it to dynamically update.\\
In Flexible ReLU \cite{46} and General ReLU, mean shift parameters were introduced to shift the mean activation close to zero. The Dual Line activation function can be viewed as a combination of DP ReLU with learn-able mean shift parameter. The better performance of Dual Line compared to DP ReLU clearly indicates the impact of the mean shift parameter.

The rest of the paper is organized as follows. Section 2 describes our proposed activation function and section 3 deals with their properties. Section 4 describes the steps to be carried out to extend the proposed concept to other activation functions. Experimental analysis and performance evaluation are described in section 5 and section 6 respectively. Results and discussion are detailed in section 7 and 8 respectively. The paper is concluded in section 9.

\section{Proposed Activation Functions}
\subsection{Dual Parametric ReLU (DP ReLU)}

DP ReLU is a new variant of ReLU with a learn-able slope parameter in both axes. The difference between Parametric ReLU (PReLU) and DP ReLU is the usage of learn-able slope parameter in the positive axis. For slope parameters ($\alpha$ and $\beta$), DP ReLU is defined as 

\begin{equation}\label{my_first_eqn}
X=
\begin{cases}
\alpha * x, & \text{if}\ x < 0 \\
\beta * x , & \text{if}\ x > 0
\end{cases}
\end{equation}

The negative slope parameter $\alpha$ is initialized with a value of 0.01 as in PReLU and Leaky ReLU. The positive slope parameter $\beta$ is initialized with a value of 1. 

\subsection{Dual Line}
Dual Line is an extension of the DP ReLU activation function. Learn-able slope parameters are multiplied to both axes and the mean shift parameter is added. The resultant activation function resembles the line equation in both axes. For slope parameters ($\alpha$ and $\beta$) and mean shift parameter (m), Dual Line is defined as

\begin{equation}\label{my_second_eqn}
X=
\begin{cases}
\alpha * x + m , & \text{if}\ x < 0 \\
\beta * x + m , & \text{if}\ x > 0
\end{cases}
\end{equation}

Mean parameter is initialized with a value of -0.22 by adding the mean shift (-0.25) and threshold (0.03) parameters used in TReLU\cite{45}. 

\pgfmathsetmacro{\first}{6}
\begin{figure}
 \subfloat[DP ReLU Activation Function \label{subfig-1:fig1}]{%
 \includegraphics[width=\first cm]{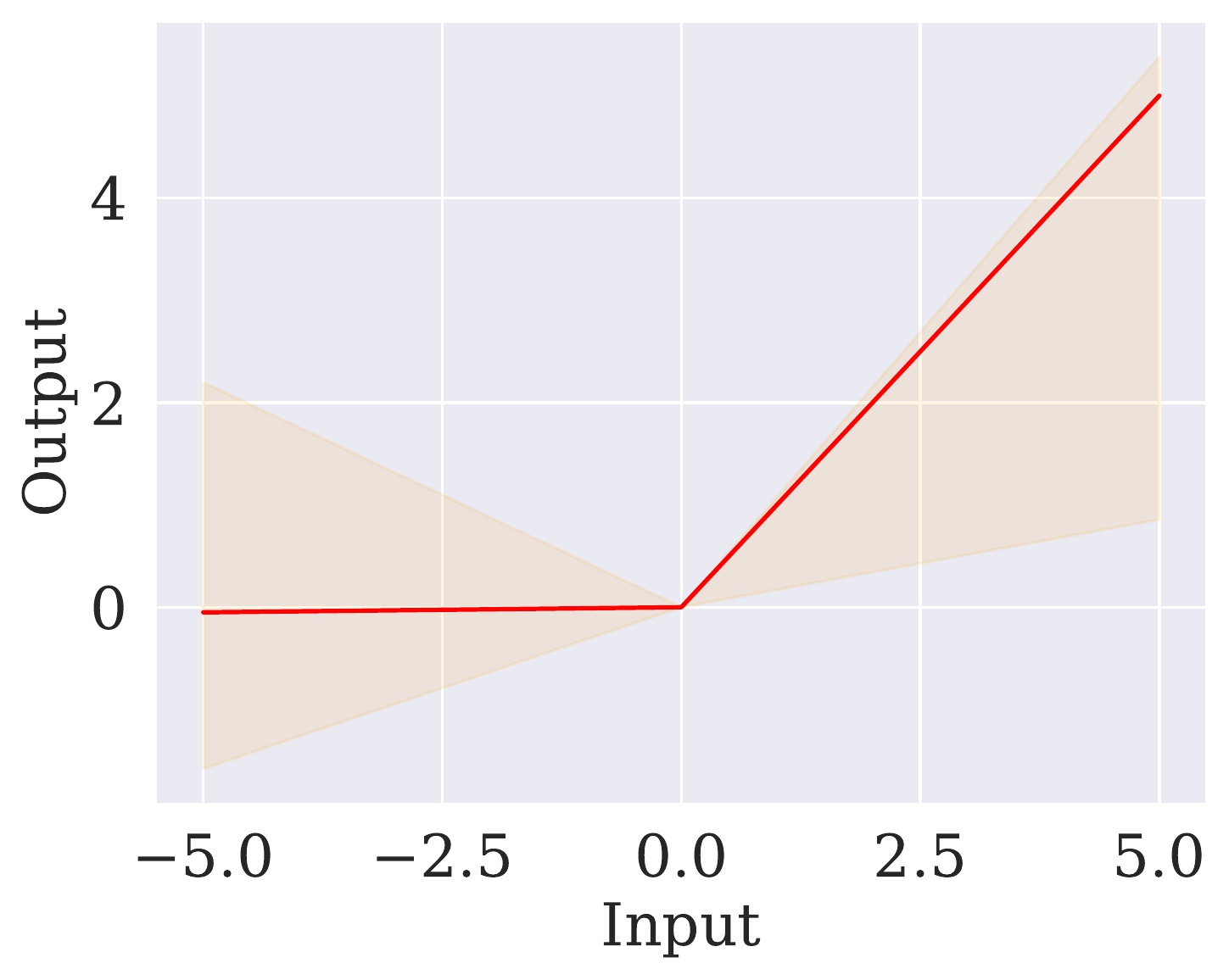}
 }
\subfloat[Dual line Activation Function \label{subfig-2:fig2}]{%
 \includegraphics[width=\first cm]{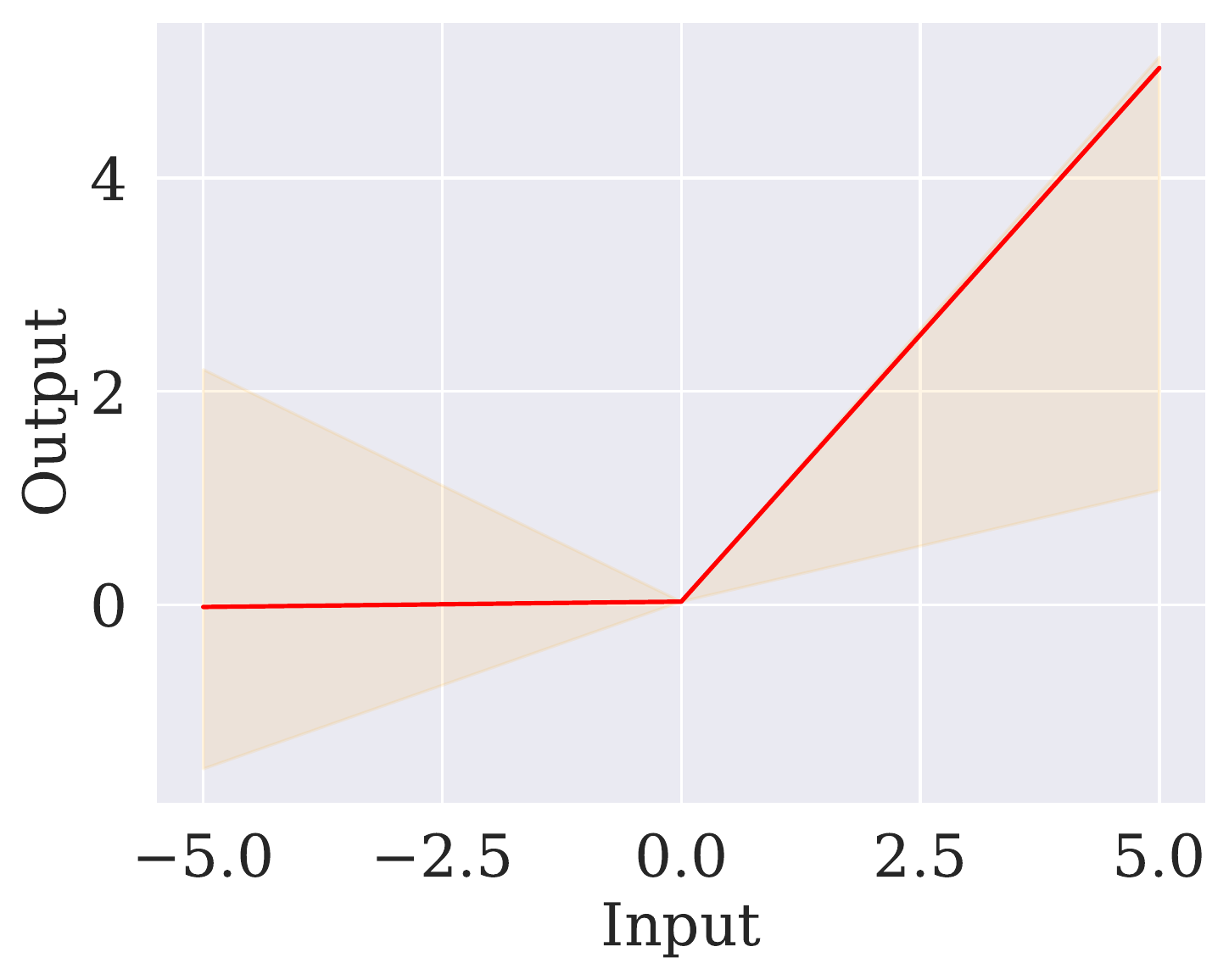}
 }
 \caption{Plot of DP ReLU and Dual Line activation function for different values of learn-able parameter. The values were obtained from WideResNet models trained on CIFAR10. Red line indicates the default initialization state. The filled region indicates the overall response region of the activation function, which is obtained by finding the min and max response curves observed across the network for the activation function.}
 
 \label{fig:dummy}
\end{figure}

\section{Properties of DP ReLU and Dual Line}
\subsection{Independent}
Both the slope parameters are independent of other parameters and act directly on the input without any constraints as shown in Equ. (\ref{my_first_eqn}) and Equ. (\ref{my_second_eqn}).

\subsection{Large response region}
As shown in Fig. 1, the learn-able parameters can take different values, so the proposed activation function has a larger response region compared to the variants without learn-able parameters. 

\subsection{Slope parameter in positive axis}
The value of $\beta$ > x results in boosting the activation and $\beta$ < x results in attenuation of the activation. The final value of $\beta$ in the model depends on the position of the activation function with respect to other layers. 

\subsection{Mean shifting due to mean shift parameter}
As per Fisher optimal learning criteria, we can reduce the undesired bias shift effect by centering the activation at zero or by using activation with negative values \cite{38}. Unit natural gradient can be achieved by pushing mean activation close to zero. This reduces the risk of over-fitting and allows the network to learn faster \cite{4},\cite{3}. The mean shift parameter in Dual Line activation function aids to push the mean activation towards zero.

\subsection{Computation requirements}
Using complex mathematical operations in the activation function increases compute time and memory requirement for training and inference. The absence of exponential or division makes ReLU and its variants faster\cite{1}. ISRLU uses inverse square roots instead of exponentials as they exhibit 1.2X better performance in Intel Xeon E5-2699 v3 (Haswell AVX2) \cite{4}. The proposed activation functions does not have complex mathematical operations. The only bottleneck in compute time during training is due to the inclusion of learn-able parameters.

\section{Extending the concept to other activation functions}
Most activation functions have an unbounded near-linear response in the Ist quadrant. The concept of adding learn-able parameter in the positive axis and mean shift parameter can be extended to other activation functions.

An existing activation function (G) can be modified by treating it as a piecewise function and replacing the characteristics for x > 0. For value of x > 0, the function can be defined as input multiplied by learn-able slope parameter and it remains the same elsewhere.

For an activation function G defined as follows,

\begin{equation}
X= G(x)
\end{equation}

The proposed concept can be applied as follows

\begin{equation}
X=
\begin{cases}
G(x) + m , & \text{if}\ x < 0 \\
\beta * x + m , & \text{if}\ x > 0
\end{cases}
\end{equation}

\section{Experimental analysis}
\subsection{Data Analysis} 
MNIST \cite{21}, Fashion MNIST \cite{27}, CIFAR10 \cite{22}, CIFAR100 \cite{22}, ImageNet 2012 \cite{23}, Tiny ImageNet \cite{24}, LFW \cite{25}, SVHN \cite{28}, and NDSB \cite{29} are the computer vision datasets used for analyzing activation functions. We are carrying out experiments with MNIST and CIFAR10 in this paper, as they are the most frequently used.

\subsection{Experimental setup}
PyTorch deep-learning library was used for the experiments \cite{30}. Adam optimizer and Flattened cross-entropy loss are used. Learning rate was estimated using learning rate finder \cite{31}. The max and min values of learning rate across multiple runs are presented, which can be an indicator for the range of values the model prefers.

With hyper-parameters and all other layers kept constant, we modified the activation function to analyze whether the activation function aids the network to learn faster and achieve better accuracy in minimal epochs. For each activation function, we ran five iterations on each of the datasets.
Computational speedup required for analyzing 43 activation functions was achieved using mixed-precision training \cite{32}.

\subsection{MNIST - LENET5 and LENET4}
LENET5 comprises of 2 convolutional layers followed by 3 linear layers. LENET4 comprises of 2 convolutional layers followed by 2 linear layers. Both LENET networks do not have batch normalization layers.

\subsection{CIFAR10 - WideResNet}
The hyper-parameters were based on fast.ai submission for the DAWNBench challenge \cite{34}. The normalized data were flipped, and random padding was carried out. The training was carried out with 512 as batch size for 24 epochs and learning rate estimated as per the learning rate estimator.
Mixup data augmentation was carried out on the data \cite{33}.

\section{Performance evaluation}
Metrics such as accuracy, top-5 accuracy, validation loss, training loss and time are estimated for the 3 networks. WideResNet contains batch norm layers and the LENET network does not. The impact of batch normalization layer would be one of the factors to consider between these architectures. The main analysis parameter is mean accuracy across 5 runs.

\section{Results}
The following sections discuss the results and analysis of training LENET5 and LENET4 on the MNIST dataset and WideResNet on the CIFAR10 dataset.

\subsection{MNIST LENET5} 
Dual Line achieves the 2nd best accuracy and best mean accuracy. DP ReLU achieves 18th and 19th in accuracy and mean accuracy respectively. Top accuracy is observed in GELU \cite{48}.\\
Dual Line achieves 2nd and 3rd rank w.r.t mean train and validation loss. DP ReLU achieves 18th and 20th in mean train and validation loss respectively. 

\pgfmathsetmacro{\first}{12}
\begin{figure}
\centering
\includegraphics[width=\first cm]{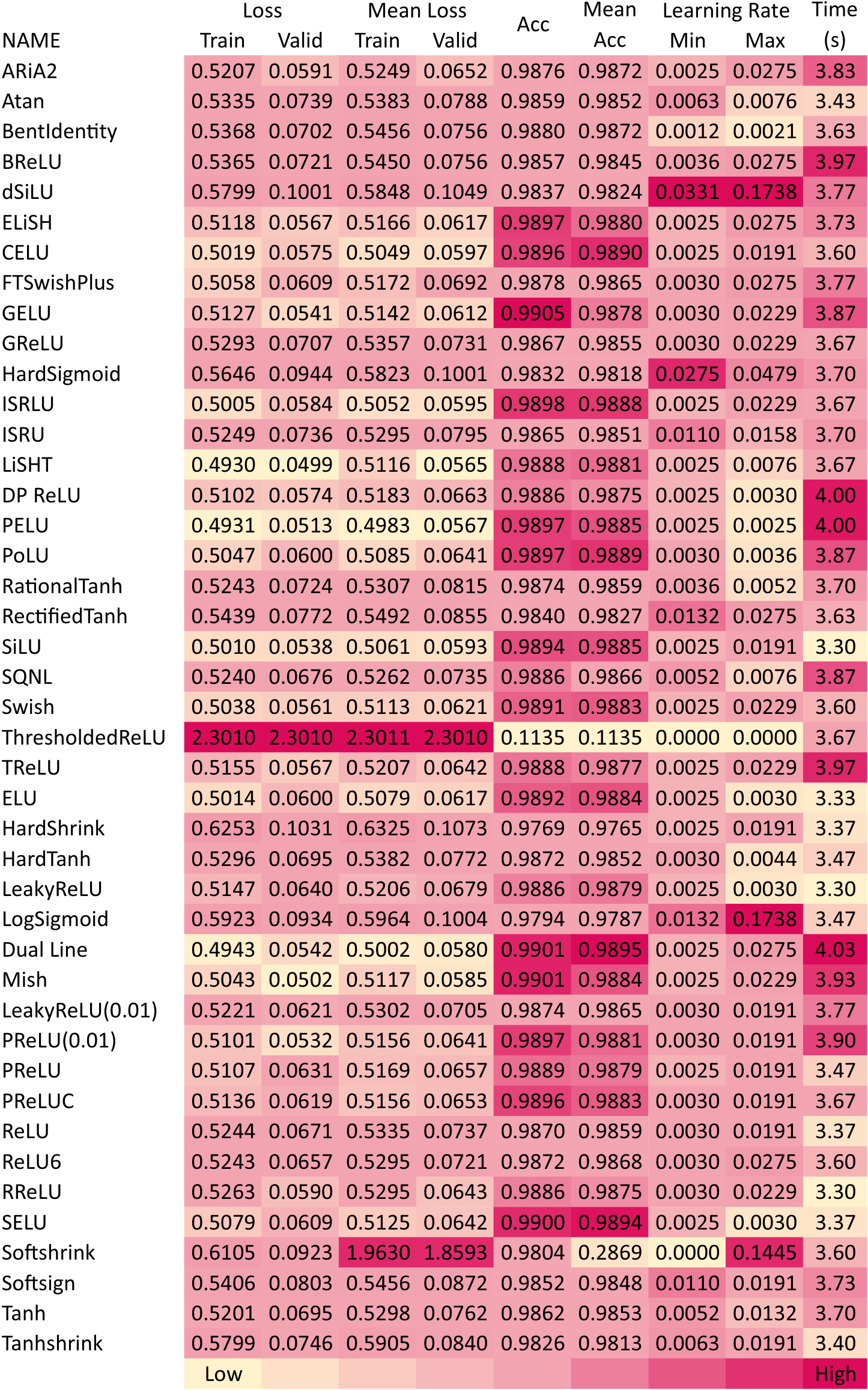}
\caption{Results for LENET5 network with different activation functions trained on the MNIST dataset. The lite to dark transition corresponds to low to high values. For loss and time, low values are preferred. For accuracy, high values are preferred. Time refers to average training time per epoch in seconds. \label{fig:Results_lenet5}}
\end{figure}

\pgfmathsetmacro{\first}{12}
\begin{figure}
\centering
\includegraphics[width=\first cm]{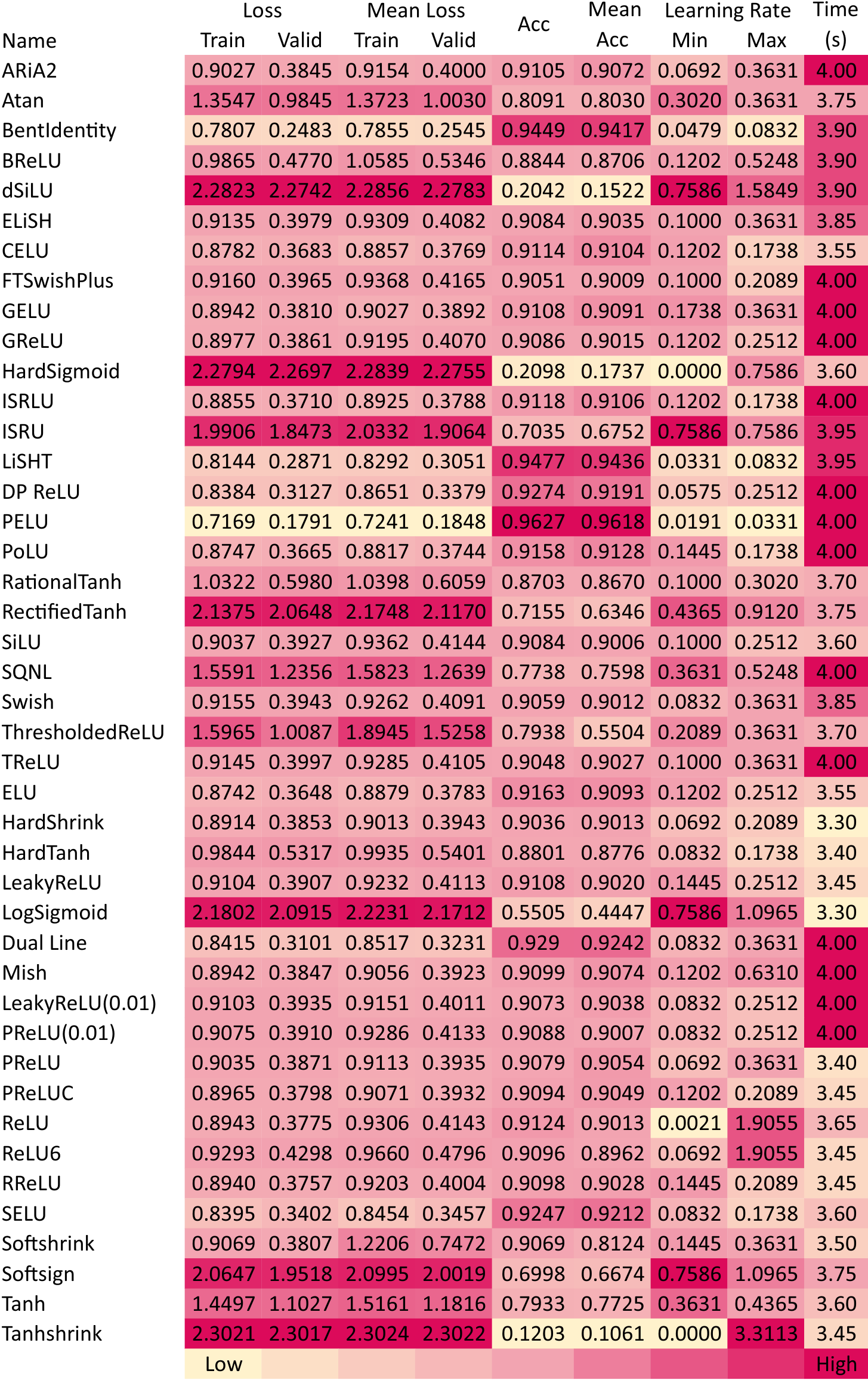}
\caption{Results for LENET4 network with different activation functions trained on the MNIST dataset. The lite to dark transition corresponds to low to high values. For loss and time, low values are preferred. For accuracy, high values are preferred. Time refers to average training time per epoch in seconds. \label{fig:Results_lenet4}}
\end{figure}

\pgfmathsetmacro{\first}{12}
\begin{figure}
\centering
\includegraphics[width=\first cm]{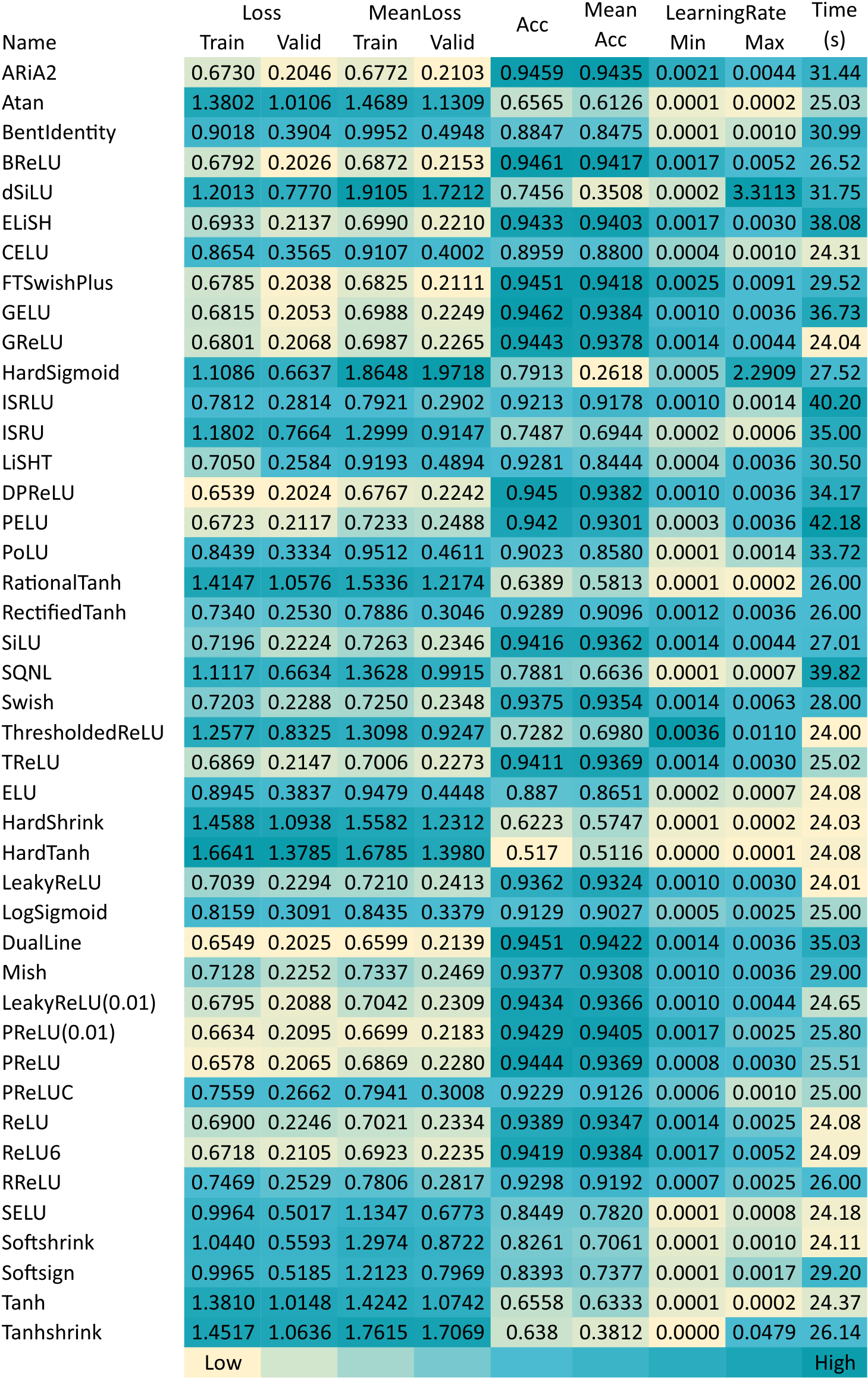}
\caption{Results for WideResNet with different activation functions trained on the CIFAR10 dataset. The lite to dark transition corresponds to low to high values. For loss and time, low values are preferred. For accuracy, high values are preferred. Time refers to average training time per epoch in seconds.
\label{fig:Results_wideresnet}}
\end{figure}

\subsection{MNIST LENET4} 
Dual Line secure 4th and 5th rank w.r.t accuracy and mean accuracy. DP ReLU secures 5th and 6th position in accuracy and mean accuracy. PELU achieves the best performance in each of the metrics.\\
Dual Line achieves 5th and 4th in mean train and validation loss. DP ReLU secures 6th and 5th in mean train and mean validation loss.

\subsection{CIFAR10 WideResNet} 
DP ReLU and Dual Line achieve 9th and 2nd best mean accuracy. Aria2 achieves the best mean accuracy of 0.9435. The highest accuracy value was observed with GELU. Dual Line and DP ReLU achieve 4th and 6th best accuracy. Best mean top-5 accuracy is observed in General ReLU. Dual Line and DP ReLU achieve 5th and 15th in mean top-5 accuracy. \\ 
DP ReLU and Dual Line achieve the best and 2nd best in Train and Validation loss. Dual Line and DP ReLU secure best and 3rd best mean train loss. 3rd and 8th best in validation loss for Dual Line and DP ReLU respectively.\\
DP ReLU performance decrease as we start to increase the number of linear layers in the model, which can be seen by comparing its performance in LENET4 and LENET5 models.

\section{Discussion}
\subsection{Learning Rate Analysis}
The learning rate estimated varies w.r.t activation used. Learning rates were estimated to indicate the range of values an activation function prefers for a dataset and are shown in the Fig. \ref{fig:Results_lenet5}, Fig. \ref{fig:Results_lenet4} and Fig. \ref{fig:Results_wideresnet}.

\setlength{\tabcolsep}{3pt}
\renewcommand{\arraystretch}{1.5}
\begin{table}[h]
\vspace{-15pt}
\caption{Analysis of learning rate values observed for each of the datasets across 5 runs. The 'Overall' row indicates the overall maximum and minimum value observed for a dataset and name of the corresponding activation function.}
\label{Table:Table_1}
\vspace{5pt}
\begin{tabular}{ccccccc}
\hline
Activation & \multicolumn{2}{c}{LENET5} & \multicolumn{2}{c}{LENET4} & \multicolumn{2}{c}{WideResNet} \\
Function & Min & Max & Min & Max & Min & Max \\
\hline
DP Relu & 2.5E-03 & 3.0E-03 & 5.7E-02 & 2.5E-01 & 1.0E-03 & 3.6E-03 \\
Dual Line & 2.5E-03 & 2.7E-02 & 8.3E-02 & 3.6E-01 & 1.4E-03 & 3.6E-03 \\
\multirow{2}{*}{Overall} & 1.0E-06 & 1.7E-01 & 1.0E-06 & 3.3E+00 & 1.0E-06 & 3.3E+00 \\
 & ThresholdedReLU & LogSigmoid & Tanshrink & Tanshrink & Tanshrink & dSiLU \\
\hline
\vspace{-20pt}
\end{tabular}
\end{table}

Table \ref{Table:Table_1} shows the maximum and minimum values observed for each of the datasets. Overall, Dual Line prefers the range as 1.4 E-03 to 0.363 and DP ReLU prefers 1E-03 to 0.251. Higher learning rates are observed in LENET4 compared to WideResNet and LENET5.

\subsection{Parameter Value Analysis}
The value of the learn-able parameters used in the activation function decides the response region, thereby the characteristics of the activation function. The neighboring blocks have an impact on the activation function as it receives input from them. The activation function requirement may vary based on the position of activation function within a block of a network, which can be analyzed by checking the parameter distribution of the activation function.
As LENET5 and LENET4 models don't have repeating blocks, we are not able to perceive any patterns as shown in Fig. \ref{fig:Value_lenet_5} and Fig. \ref{fig:Value_lenet_4}.

\pgfmathsetmacro{\first}{6}
\vspace{-15pt}
\begin{figure}
 \subfloat[DP ReLU activation function \label{subfig-3:fig1}]{%
 \includegraphics[width=\first cm]{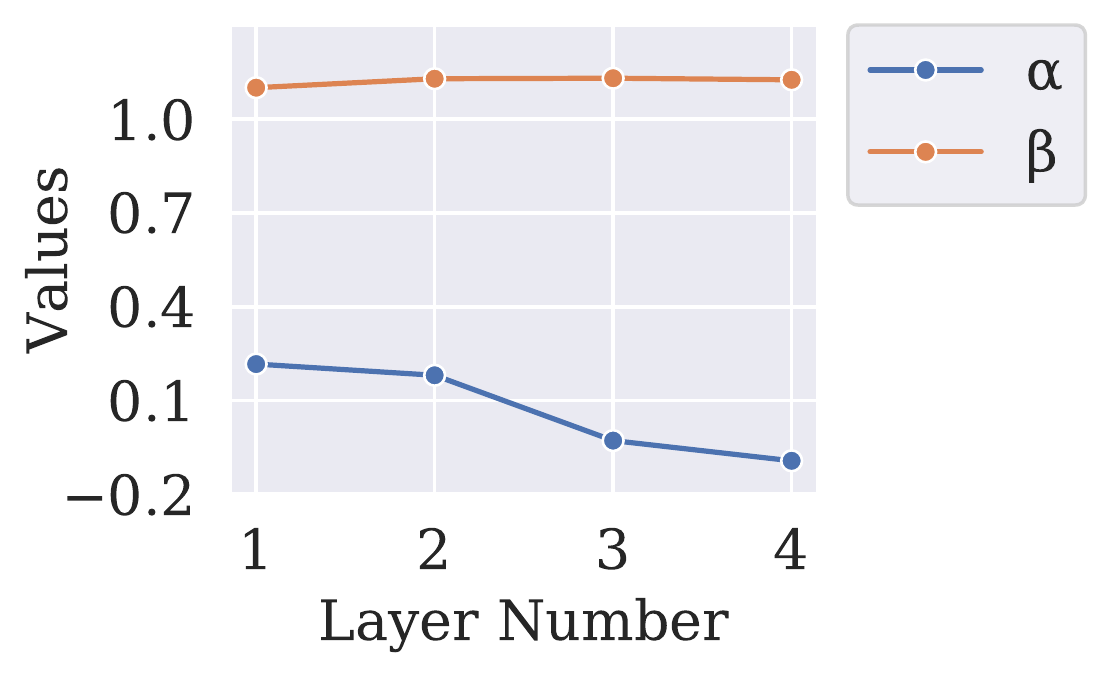}
 }
\subfloat[Dual Line activation function \label{subfig-4:fig2}]{%
 \includegraphics[width=\first cm]{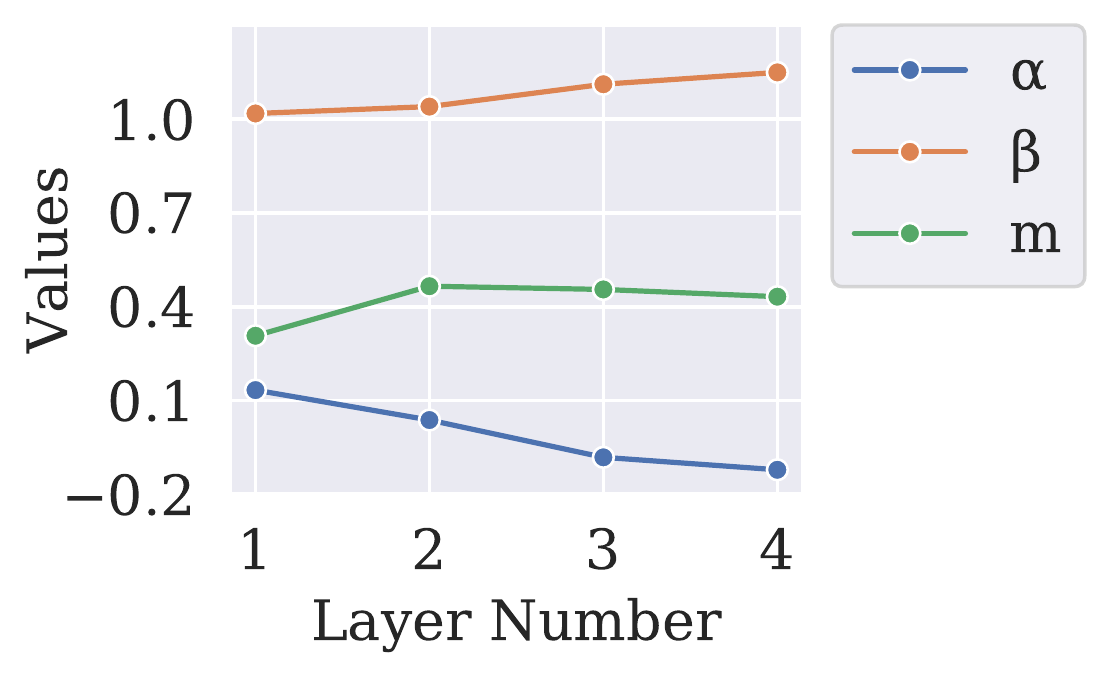}
 }
 
 \caption{Value of parameters w.r.t position of the activation function from top to bottom of LENET5 model }
\label{fig:Value_lenet_5}
\end{figure}

\pgfmathsetmacro{\first}{6}
 \vspace{-40pt}
\begin{figure}
 \subfloat[DP ReLU activation function\label{subfig-5:fig1}]{%
 \includegraphics[width=\first cm]{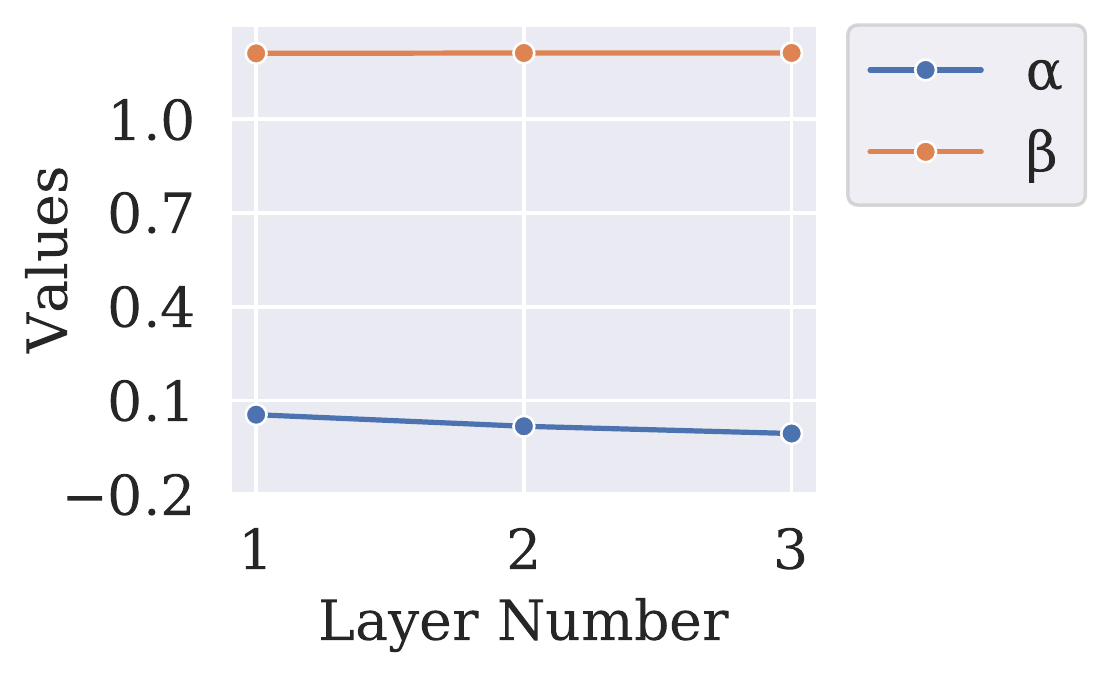}
 }
\subfloat[Dual Line activation function\label{subfig-6:fig2}]{%
 \includegraphics[width=\first cm]{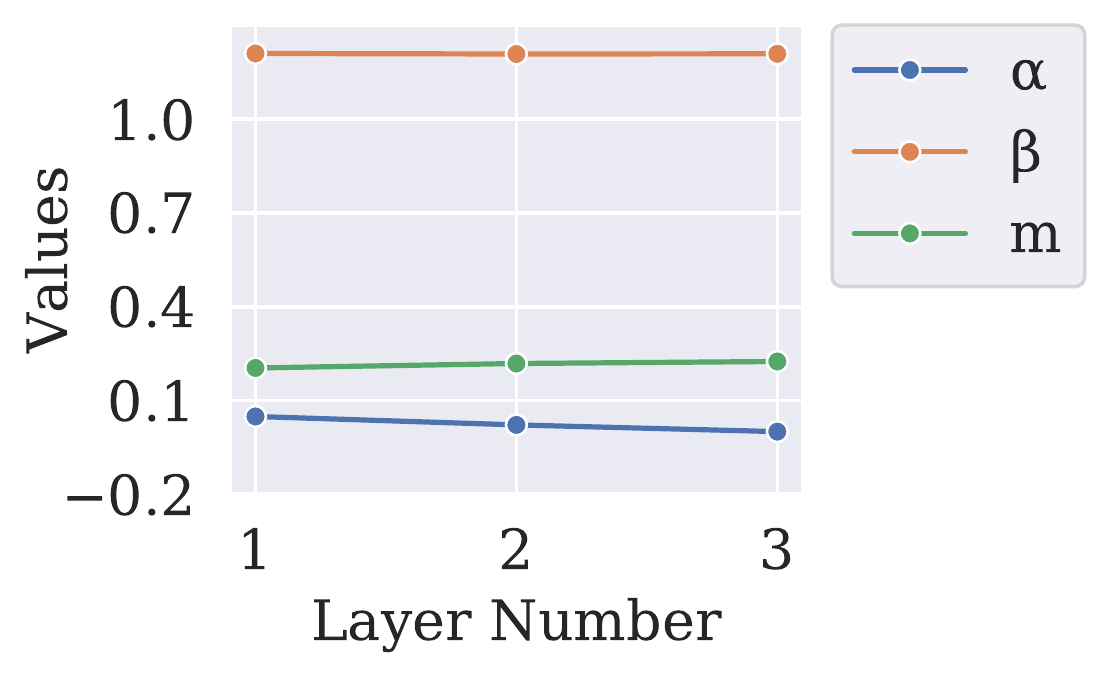}
 }
 \caption{Value of parameters w.r.t position of the activation function
from top to bottom of LENET4 model}
 \vspace{-15pt}
 \label{fig:Value_lenet_4}
\end{figure}

In WideResNet, activation function occurs twice in each of the nine repeating blocks of the network. The distribution of parameter values w.r.t each of these blocks is analyzed to view the relationship existing between activation functions occurring within a same block, as shown in Fig. \ref{subfig-2:Block_analysis_fig1} and Fig. \ref{subfig-2:Block_analysis_fig2}.
The value of $\beta$ of the 1st activation function within a block is larger than the 2nd activation function in the blocks( B1-B8). The first block (B0) which is close to the input, does not exhibit this pattern.

The box plot of $\alpha$ value indicates the marginal difference in distribution within a block as shown in Fig. \ref{subfig-1:Box_plot_dp_relu_fig1}. The distribution of $\beta$ value differs a lot within a block as shown in Fig. \ref{subfig-2:Box_plot_dual_line_fig2}. The outliers present in the box plot are due to the $\beta$ values from block (B0). This indicates that the activation function requirements differ within a block of a network.

\pgfmathsetmacro{\first}{6}
\begin{figure}
\vspace{-25pt}
 \subfloat[Distribution of $\alpha$ parameter \label{subfig-1:Box_plot_dp_relu_fig1}]{%
 \includegraphics[width=\first cm]{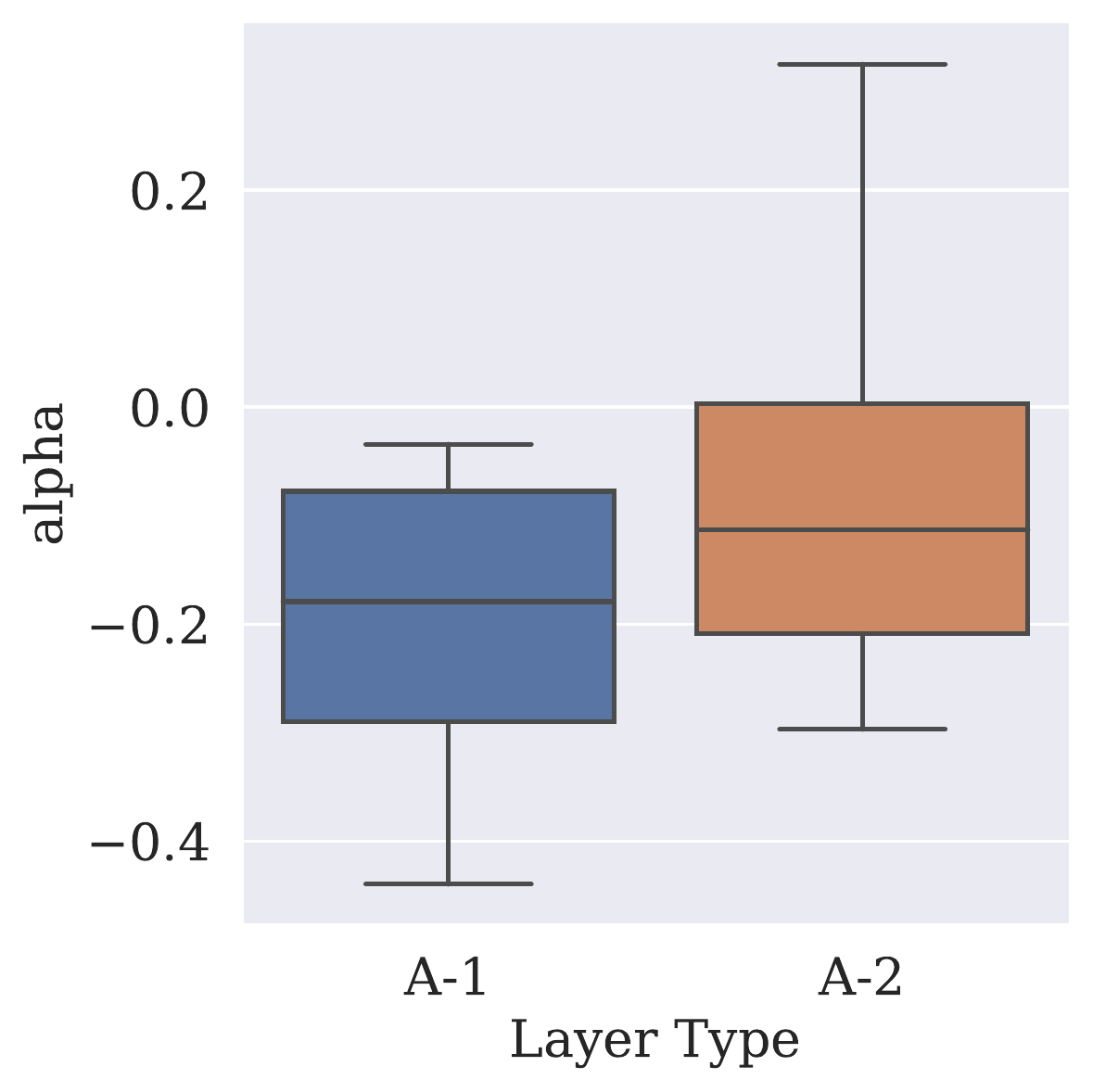}
 }
\subfloat[Distribution of $\beta$ parameter \label{subfig-2:Box_plot_dp_relu_fig2}]{%
 \includegraphics[width=\first cm]{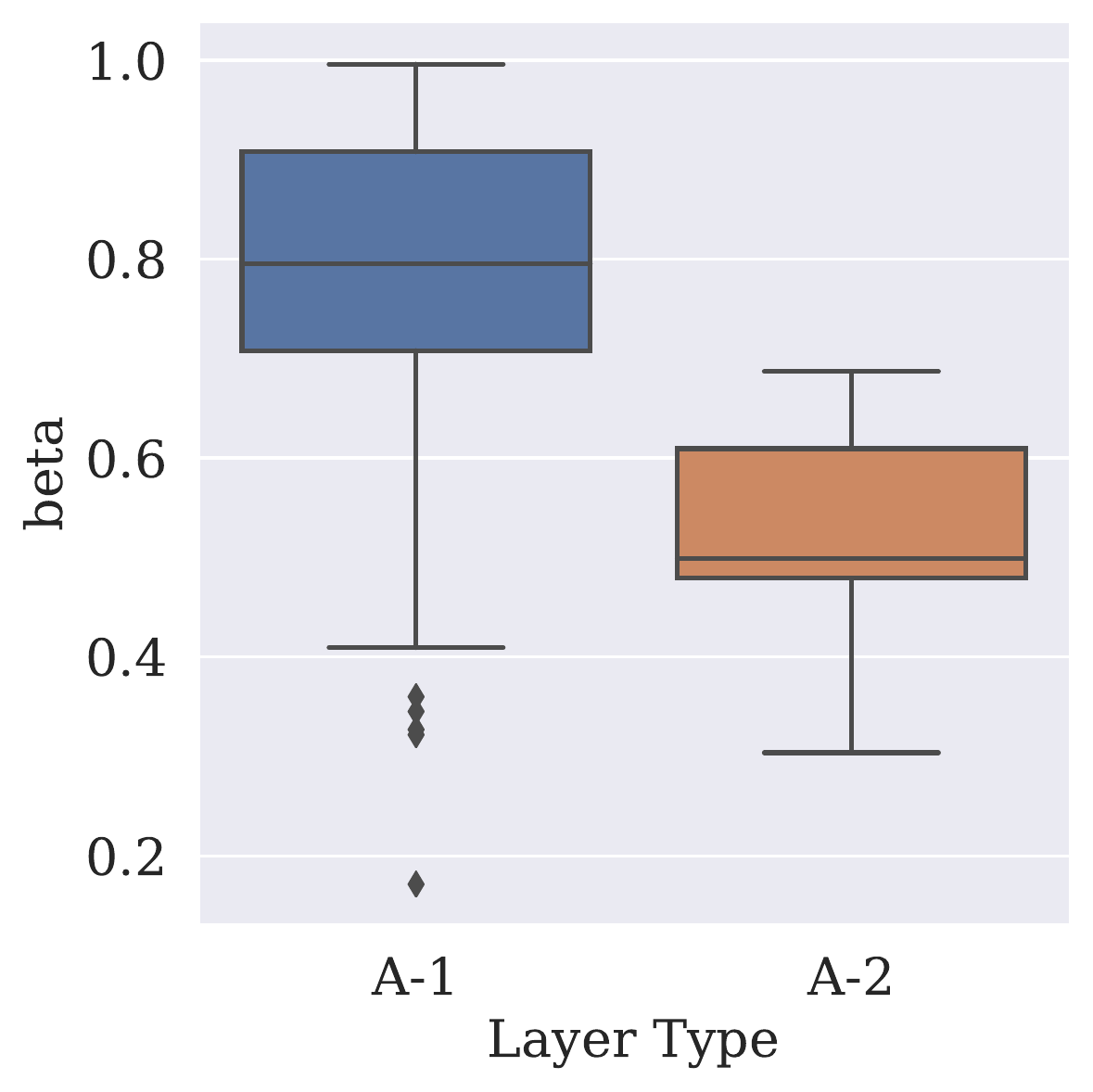}
 }
 \caption{Box plot of parameters of DP ReLU activation within each block. A-1 and A-2 represent the first and second activation function within each block of WideResNet.}
 \vspace{-25pt}
 \label{fig:Box_plot_dp_relu}
\end{figure}

\pgfmathsetmacro{\first}{6}
\begin{figure}
 \vspace{-15pt}
 \subfloat[Distribution of $\alpha$ parameter \label{subfig-1:Box_plot_dual_line_fig1}]{%
 \includegraphics[width=\first cm]{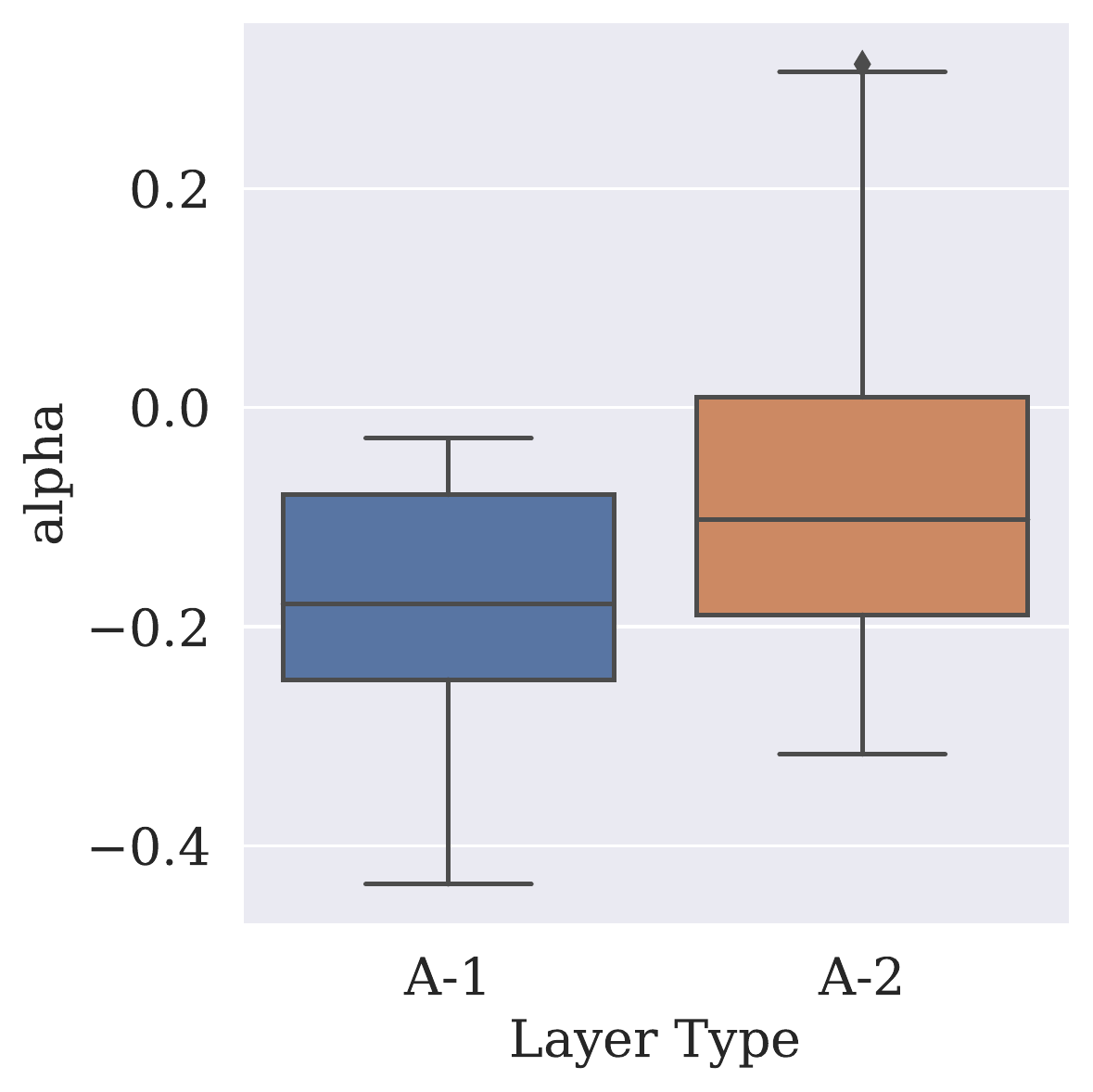}
 }
\subfloat[Distribution of $\beta$ parameter \label{subfig-2:Box_plot_dual_line_fig2}]{%
 \includegraphics[width=\first cm]{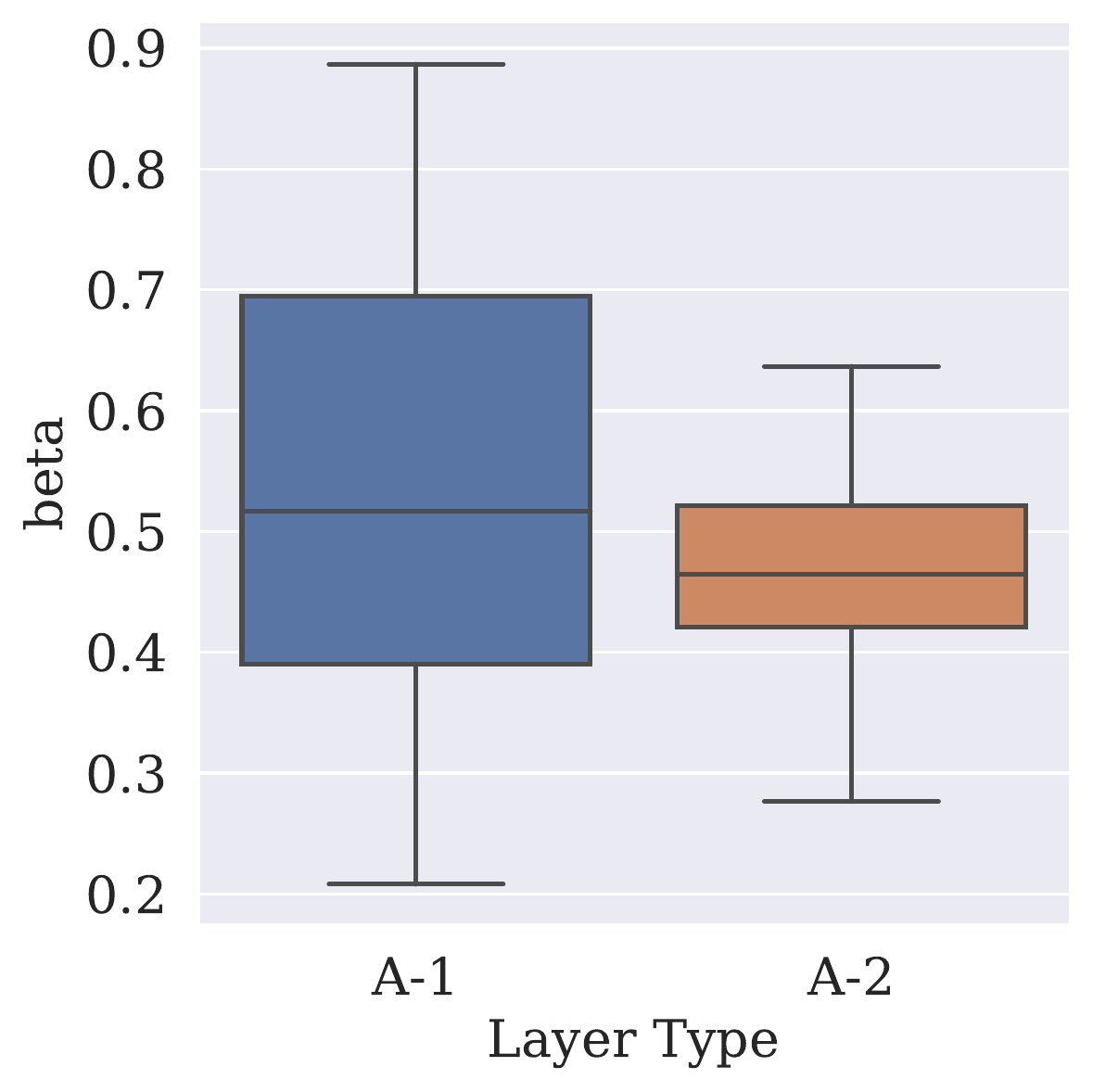}
 }
 \caption{Box plot of parameters of Dual Line activation within each block. A-1 and A-2 represent the first and second activation function within blocks of WideResNet.}
 \vspace{-15pt}
 \label{fig:Box_plot_dual_line}
\end{figure}

\pgfmathsetmacro{\first}{11}
\begin{figure}[!htbp]
\begin{minipage}{\linewidth}
\subfloat[ Value of $\alpha$ and $\beta$ parameters of DP ReLU trained on CIFAR10 \label{subfig-2:Block_analysis_fig1}]{%
 \includegraphics[scale=0.7]{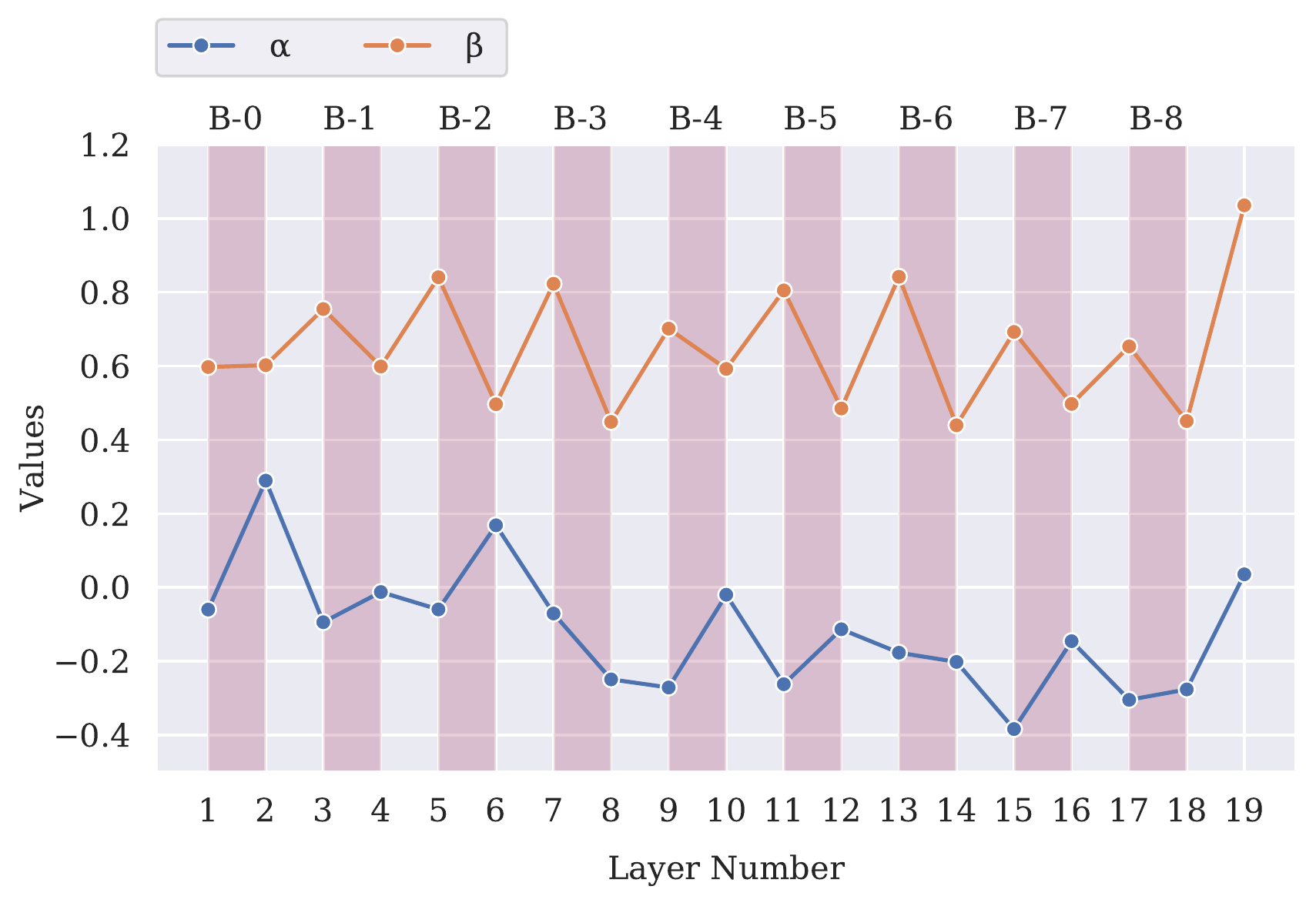}
}
\end{minipage}

\begin{minipage}{\linewidth}
\subfloat[Value of $\alpha$, $\beta$ and mean shift parameters of Dual Line trained on CIFAR10 \label{subfig-2:Block_analysis_fig2}]{%
 \includegraphics[scale=0.7]{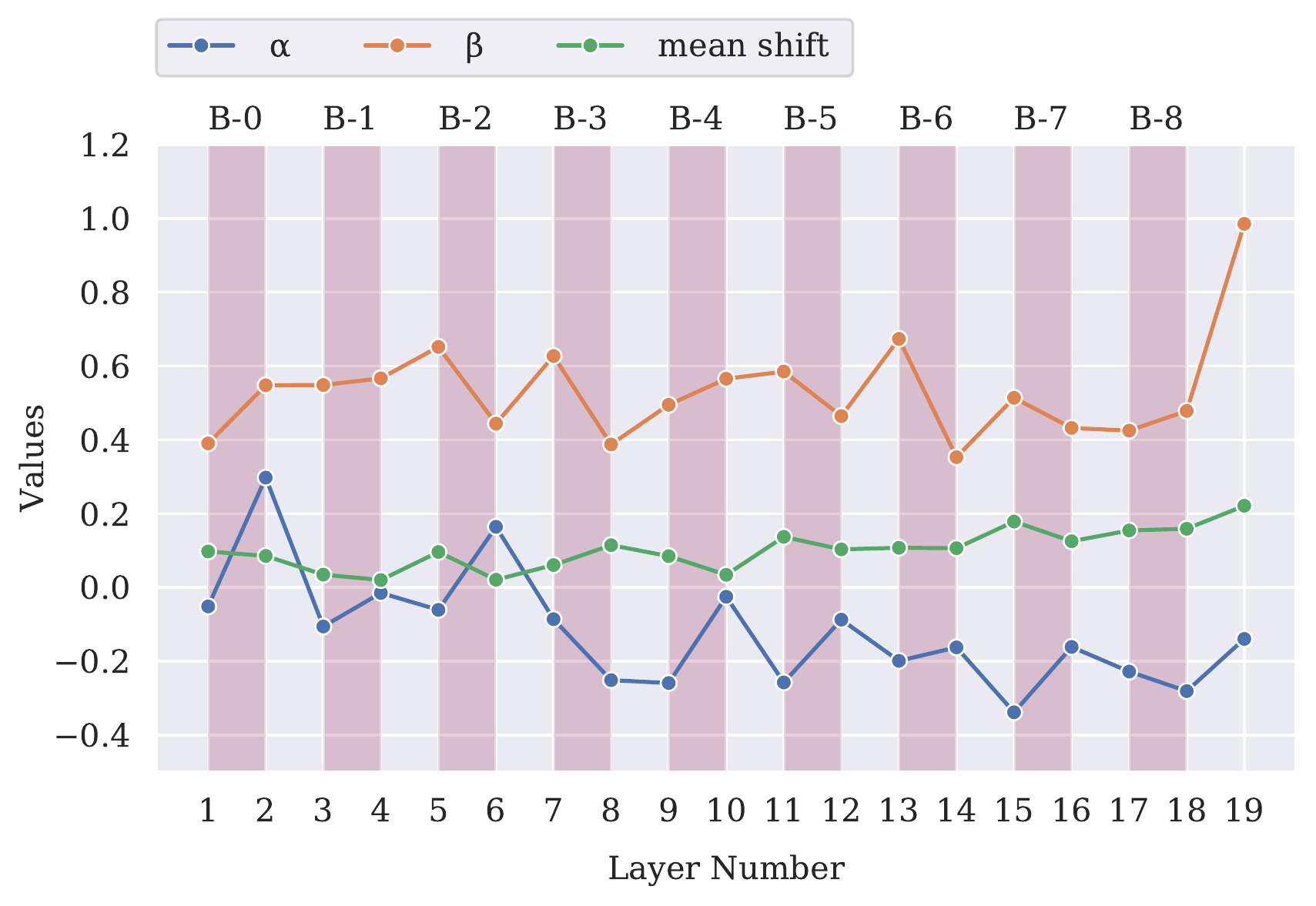}
}
\end{minipage}

\caption{Distribution of parameter values for proposed activation functions across blocks of WideResNet. X - axes represent the position of the activation function from top to bottom of the network. Red vertical blocks (B-0 to B-8) represent each of the blocks.}
\label{fig.Block_analysis}

\end{figure}

From Fig. \ref{subfig-2:Block_analysis_fig2}, we can observe that the $\alpha$ parameter does not exhibit any significant pattern, the mean shift parameter is nearly constant and $\beta$ values differ marginally for Dual Line. The marginal difference in $\beta$ distribution can be seen clearly in Fig. \ref{subfig-2:Box_plot_dual_line_fig2}.
The block-level parameter distribution pattern indicates the final distribution of the parameters and helps us to understand the requirements of activation function at block level. As we have used learn-able parameters, the proposed activation function can adapt as per the requirement posed by its neighboring layers. 

\subsection{Performance on fast.ai Leaderboards}
The proposed activation function Dual Line broke 3 out of 4 fast.ai leaderboards for the image classification task. It was able to achieve more than 2\% percent improvement in accuracy in two leaderboards. Parameter distribution pattern within blocks was observed in XResNet-50 models trained for the above 4 leaderboards.\\
In our current work, we have analyzed the concept of learn-able slope parameter for the positive axis and mean shift parameter and have shown the performance benefit of the same. Our future work deals with reducing the computation time taken during training.

\section{Conclusion}
The novel concept of adding a learn-able slope and mean shift parameter is introduced in this paper. Overall, our experiments indicate the performance benefit of the proposed concept. The concept can be added to other activation functions with ease for performance boost. As the paper captures the activation function requirement at the block level, the proposed concept can be used as a supporting guideline for developing new activation functions for computer vision.

\bibliography{main_paper}
\bibliographystyle{splncs04}

\end{document}